\documentclass[conference]{IEEEtran}
\IEEEoverridecommandlockouts
\usepackage{cite}
\usepackage[numbers]{natbib}
\usepackage{amsmath,amssymb,amsfonts}
\usepackage{algorithmic}
\usepackage{graphicx}
\usepackage{textcomp}
\usepackage{xcolor}
\usepackage[binary-units=true]{siunitx}
\usepackage{subcaption}
\usepackage{booktabs}
\usepackage[bookmarks=false]{hyperref}
\usepackage{cancel}
\usepackage{mathtools}
\def\BibTeX{{\rm B\kern-.05em{\sc i\kern-.025em b}\kern-.08em
    T\kern-.1667em\lower.7ex\hbox{E}\kern-.125emX}}

\DeclareMathOperator{\prob}{P}
\newcommand{\uprob}[1]{\prob\!\left(#1\right)}
\newcommand{\cprob}[2]{\prob\!\left(#1 \middle| #2\right)}
    
\begin{document}
\title{Learning to Predict Robot Keypoints Using Artificially Generated Images}

\newif\iffinalcopy

\finalcopytrue

\iffinalcopy
    \author{
        \IEEEauthorblockN{Christoph Heindl}
        \IEEEauthorblockA{
            \textit{Visual Computing Department}\\
            PROFACTOR GmbH \\ 
            4407 Steyr, Austria\\ 
            \texttt{cheind@profactor.at}
        }
        \and
        \IEEEauthorblockN{Sebastian Zambal}
        \IEEEauthorblockA{
            \textit{Machine Vision Department}\\
            PROFACTOR GmbH \\ 
            4407 Steyr, Austria\\ 
            \texttt{szamba@profactor.at}
        }
        \and
        \IEEEauthorblockN{Josef Scharinger}
        \IEEEauthorblockA{
            \textit{Institute of Computational Perception} \\
            Johannes Kepler University \\ 
            4040 Linz, Austria\\ 
            \texttt{josef.scharinger@jku.at}
        }
    }

\else

    \author{
        \IEEEauthorblockN{
            Anonymous
        }
    }

\fi

\maketitle

\begin{abstract}
    This work considers robot keypoint estimation on color images as a supervised machine learning task. We propose the use of probabilistically created renderings to overcome the lack of labeled real images. Rather than sampling from stationary distributions, our approach introduces a feedback mechanism that constantly adapts probability distributions according to current  training progress. Initial results show, our approach achieves near-human-level accuracy on real images. Additionally, we demonstrate that feedback leads to fewer required training steps, while maintaining the same model quality on synthetic data sets.
\end{abstract}


\section{Introduction}
In this paper, we study the task of detecting robot keypoints on 2D color images. The visual perception of robots enables machines to quickly capture and build interaction patterns, without the need for tedious spatial and temporal synchronization. Robot keypoint detection represents, therefore, a key component in human-machine and ad-hoc machine-machine collaboration.


The detection of robot joints in color images of cluttered industrial environments is a challenging task. Robot appearance varies in part due to positioning of joints, form factors, texturing, illumination, and self-occlusions. The loss of depth through perspective camera projection adds an additional layer of complexity. 

Prior work mostly deals with depth-enriched input data. These approaches frame localization as an iterative application of the alignment of point clouds and CAD parts. On the contrary, keypoint estimation on color images requires complex feature detectors, capable of identifying all the different ways in which robots appear in images. Data-driven machine-learning methods have shown that such robust patterns can be learned from training data. However, the lack of comprehensive training databases has prevented the latest results of deep learning from being applied to the task of robotic keypoint detection.


\begin{figure}[!tbp]
    \centerline{
        \iffinalcopy
            \includegraphics[width=\columnwidth]{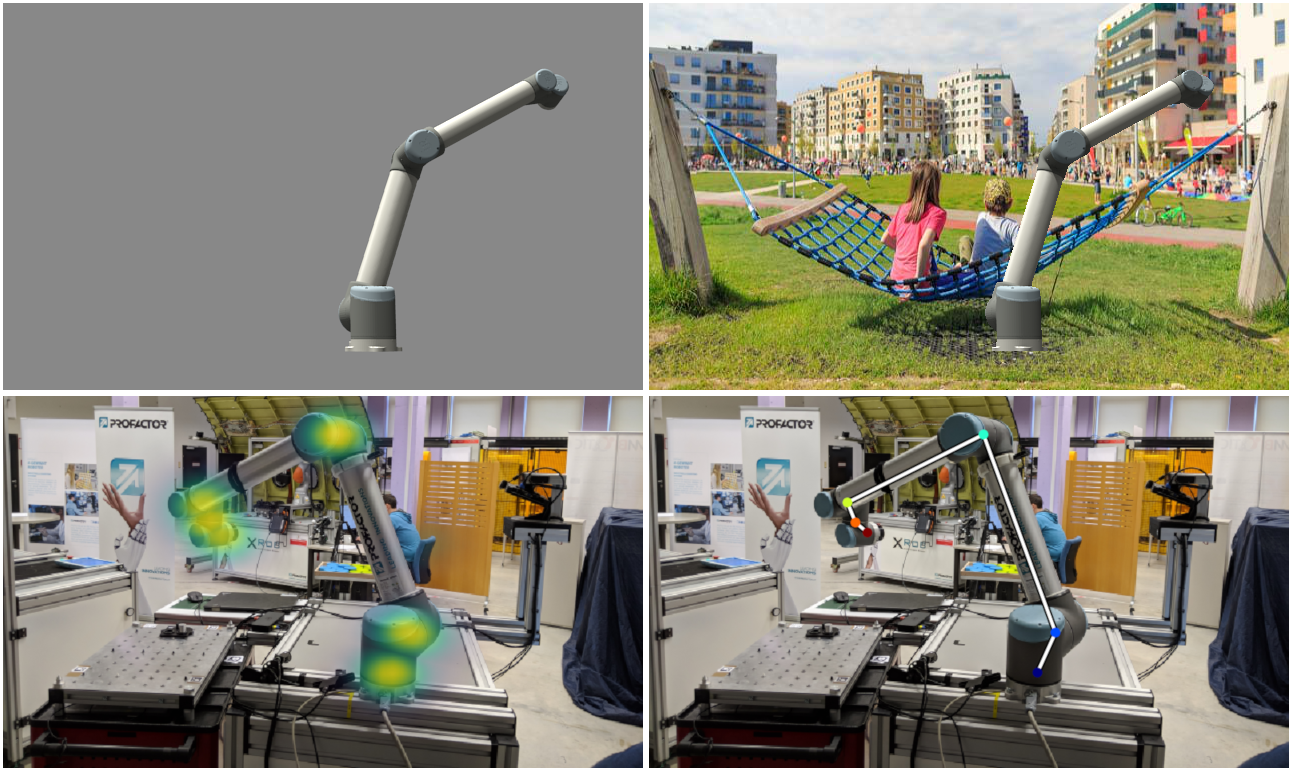}
        \else
            \includegraphics[width=\columnwidth]{figures/frontpage/renderedImages_blind.png}
        \fi
    }
    \caption{\label{fig:intro} \textbf{Top}: Our approach produces non photo-realistic training images by rendering robot scenes controlled by a probabilistic model. \textbf{Bottom left}: Robot joint belief maps as predicted by the localization network. \textbf{Bottom right}: Individual joint positions linked to build a kinematic robot chain. Best viewed in color.}
\end{figure}

In this work, we present an efficient method for detecting robot joints in color images using deep convolutional neural networks (DCNNs). As outlined in \figurename \ref{fig:intro}, our approach addresses the scarcity of labeled data by training on non-photo-realistic computer generated images. We introduce the idea of a feedback mechanism that allows information to flow from model training into the probabilistic aspects of the simulation. Based on this feedback, we optimize the governing probability distributions towards training needs. We present promising initial results on artificial and restricted real data sets. The results of a pilot study indicate that the accuracy of system predictions is comparable to human performance on real images. We have released source code for our adaptive training architecture to ensure reproducibility and promote adaptation to other areas.




\section{Related Work}
In contrast to human pose estimation from color input images, few approaches for robot pose estimation have been proposed. 

In Miseikis et al. \cite{miseikis2019two} a multi-task DCNN is proposed that learns to predict robot joint positions, robot type and segmentation masks from color input images. In contrast to this work, Miseikis et al. train on pictures of real robots, collected using a calibrated color/depth sensor. The effort for creating these training samples is considerable and limited to the depth range of the sensor used for data acquisition. The work most closely related to ours is Heindl et al. \cite{cheind2019disp}, who propose a two-layer architecture that learns joint positions and instance segmentation masks from artificial color input images. We take up their idea of artificial data generation and formulate it in terms of a probabilistic model. This way, posterior inference can be exploited to accelerate model learning.

From a data augmentation perspective, the family of Generative Adversarial Networks (GANs) \cite{goodfellow2014generative} provide a generic framework for artificial data generation from noisy input. However, application to our use case requires generated images to carry precise meta-information (e.g joint positions). Such conditional GAN approaches \cite{isola2017image} are much harder to train and require a large amount of labeled input. Because we try to avoid tedious real data acquisition, we do not consider this approach for the remainder of this work. However, we do highlight the work of Sixt et al. \cite{sixt2018rendergan} who propose the use of GANs in combination with deterministic simulators to add the necessary levels of realism to images, while guaranteeing pixel-exact semantic context. 

\section{Method}
    The following sub-sections are structured as follows. First, we describe the localization network to be trained. Next, we introduce the probabilistic model and attached rendering simulator. Finally, we introduce the main system architecture for adaptive training.

    Throughout this work, lower-case non-bold characters denote scalars. Bold-faced lower-case characters indicate random samples and higher order tensors. Upper-case (bold) characters denote (multivariate) random variables. 
    
    \subsection{Localization Model}
        \label{sec:model}
        The localization model for robot joint prediction is based on an architecture proposed by Heindl et al. \cite{cheind2019disp}, which we briefly report for clarity. 
        
\begin{figure} [!h]
            \centering
            \includegraphics[width=0.9\columnwidth]{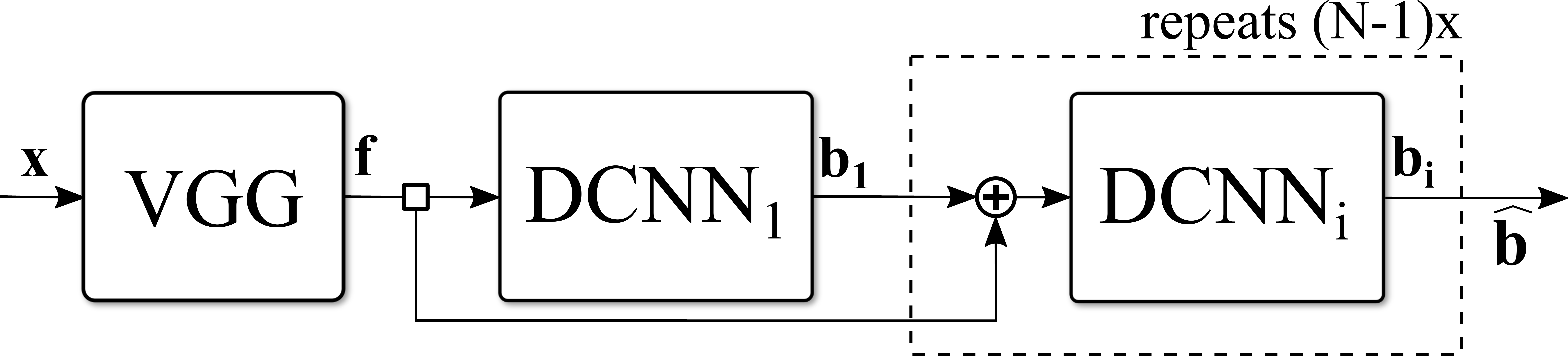}
            \caption {
                \label{fig:jointmodel} 
                Localization network based on Heindl et al. \cite{cheind2019disp}: Given input image $\mathbf{x}$, a pre-trained feature extractor (VGG) computes base features $\mathbf{f}$. The first stage $\textrm{DCNN}_1$ transforms $\mathbf{f}$ into prediction $\mathbf{b}_1$. A sequence of follow-up stages $\textrm{DCNN}_i$ iteratively refines $(\mathbf{b}_{i-1},\mathbf{f})$ to yield $\mathbf{b}_i$. The output of the last stage $\textrm{DCNN}_N$ is the model prediction $\hat{\mathbf{b}} = \mathbf{b}_N$.
            }
        \end{figure}

        The structure of the network is depicted in \figurename \ref{fig:jointmodel}. Given a color image $\mathbf{x} \in \mathbb{R}^{3 \times H\times W}$, the network outputs joint position belief maps $\hat{\mathbf{b}} \in \mathbb{R}^{J \times H \times W}$, where $J$ is the number of robot joints. Each belief map encodes the likelihood of observing the given joint at a specific pixel location. 
        The localization model first uses a pre-trained network (VGG) \cite{simonyan2014very} to perform initial feature extraction using $C$ channels $\mathbf{f} \in \mathbb{R}^{C \times H \times W}$ from $\mathbf{x}$. These features are fed into a sequence of $N$ identical DCNN blocks, i.e. stages. The $i$th stage takes $\mathbf{f}$ and $\mathbf{b}_{i-1}$ as input and outputs a refined belief map $\mathbf{b}_{i}$ via a series of convolutional layers. The output of the final stage $N$ represents the final prediction $\hat{\mathbf{b}} = \mathbf{b}_N$. Using ground truth maps $\mathbf{b}^*$, the network parameters are learned by optimizing  pixel-wise stage losses $L\colon \mathbb{R}^{J \times H \times W}, \mathbb{R}^{J \times H \times W} \to \mathbb{R}$ given by $\mathbf{b}_i, \mathbf{b}^* \mapsto L(\mathbf{b}_i, \mathbf{b}^*)$ after each stage. The total loss is the average over all stage losses.
        
        
        

    \subsection{Artificial Data Generation}
        \label{sec:artificialData}       
        We formulate data generation as a probabilistic model that interacts with deterministic rendering functions. The probabilistic aspects govern scene composition, while the deterministic functions are responsible for creating images from given scene descriptions. Our probabilistic model includes  robot joint angles, view perspective and material properties.

        \begin{figure}[htbp]
            \centerline{\includegraphics[width=0.7\columnwidth]{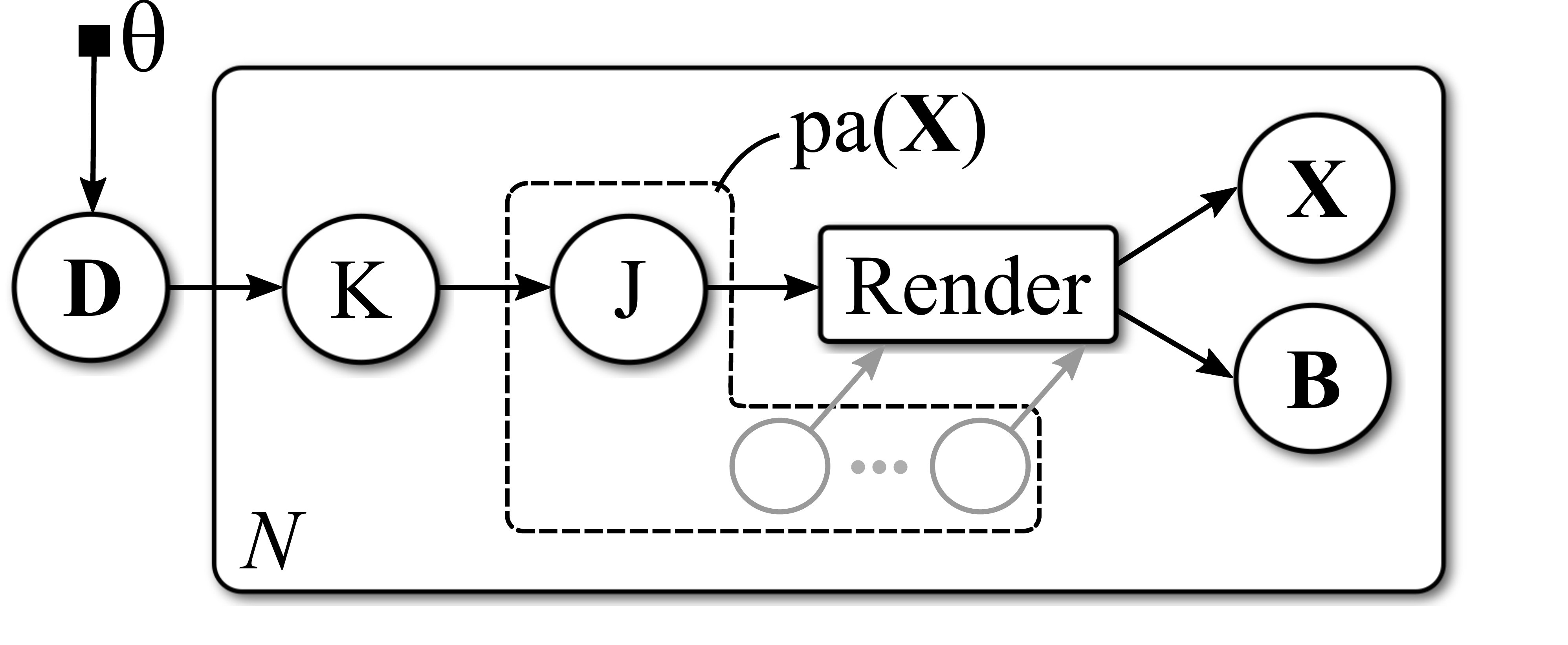}}
            \caption{\label{fig:probModel} An excerpt of our probabilistic renderer focusing on a single robot joint $J$. The created image $\textbf{X}$ and robot joint beliefs $\textbf{B}$ are determined by random samples of robot joint angle $J$ conditional on bin $K$. The distribution of $K$ itself is governed by prior $\textbf{D}$. (Gray: other random variables are not specified in detail here, ${\textrm{pa}(\textbf{X})}$ denotes the set of parents of $\textbf{X}$).}
        \end{figure}
        
        \figurename \ref{fig:probModel} depicts our data generation process, focusing on a single robot joint angle $J$. We discretize the value range $\left[0,2\pi\right]$ of $J$ into $K$ bins where the probability distribution over all bins is governed by prior $\textbf{D}$. For a given bin $k$, the joint angle $J$ is distributed uniformly within the corresponding bin range. The render function transforms $J$ (and other random samples) into training image $\textbf{X}$ and joint belief label $\textbf{B}$. Formally, we let
        \begin{align*}
            \textbf{D} &\sim \textrm{Dirichlet}(\theta)\\
            K \mid \textbf{D}\texttt{=}\textbf{d} &\sim \textrm{Categorical}(\textbf{d})\\
            J \mid K\texttt{=}k &\sim \textrm{Uniform}(\textrm{lo}(k), \textrm{hi}(k)),  
        \end{align*}
        where $\textrm{lo}(k)$, $\textrm{hi}(k)$ are the interval limits of the $k$th bin.  Decoupling the probabilistic aspects and image generation allows the use of methods of probabilistic inference to alter the system's generative behavior. In the next sub-section we detail this approach and introduce a feedback channel to adapt image generation to current training loss.

    \subsection{Adaptive Training Architecture}
        \label{sec:adaptivetraining}
\begin{figure}[htbp]
            \centerline{\includegraphics[width=0.68\columnwidth]{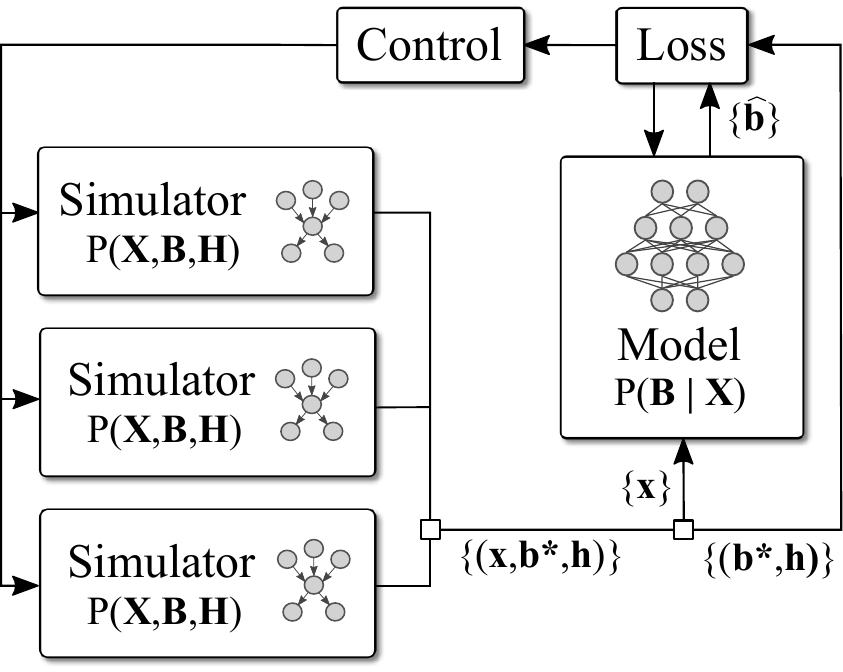}}
            \caption{\label{fig:architecture} The adaptive training architecture. Based on a probabilistic description, multiple simulator instances generate training samples $\{(\textbf{x},\textbf{b}^*,\textbf{h})\}$. Input images $\textbf{x}$ and labels $\textbf{b}^*$ are considered for neural network training. Using information from a single training step, a control block adapts the probabilistic rendering behavior to better match requirements.}
        \end{figure}

        \figurename \ref{fig:architecture} outlines our proposed training architecture. One or more simulator engines generate training tuples $\{(\textbf{x},\textbf{b}^*,\textbf{h})\} \sim \uprob{\textbf{X},\textbf{B},\textbf{H};\theta}$ by sampling from the probabilistic rendering model. Here, $\textbf{H}$ denotes all hidden variables except $\textbf{B}$. The observables $\textbf{x}$ are passed through a deep neural network parametrized by $\mathbf{\phi}$ to predict $\hat{\textbf{b}}$. The prediction $\hat{\textbf{b}}$ and supervised targets $\textbf{b}^*$ are then used for classic mini-batch updating of the network parameters $\mathbf{\phi}$ using a suitable loss function \cite{cheind2019disp}.

        A key idea to our work is a feedback mechanism that enables adaptive sampling with simulators. As outlined in \figurename \ref{fig:architecture}, a controller updates the simulation based on information of the current training state. We model these updates as inference over hidden variables given a selected set $\mathcal{S} = \{(\textbf{x},\textbf{b}^*,\textbf{h})\}$ of training samples. 
        

        We follow-up on the model proposed in sub-section \ref{sec:artificialData} to illustrate the inference for the distribution $\textbf{D}$ that governs joint $J$. The posterior distribution over $\textbf{D}$ is given by $\cprob{\textbf{D}}{\textbf{X}, K, J;\mathcal{S}}$. In particular, according to our model, the joint distribution factors as $$\uprob{J, K, \textbf{D}, \textbf{X}} = \uprob{\textbf{D}}\cprob{K}{\textbf{D}}\cprob{J}{K}\cprob{\textbf{X}}{\textrm{pa}(\textbf{X})},$$ where ${\textrm{pa}(\textbf{X})}$ refers to all direct parents of $\textbf{X}$. Deriving the posterior distribution simplifies to 
        \begin{align*}
            \cprob{\textbf{D}}{\textbf{X}, K, J} &= \frac{\uprob{\textbf{D}}\cprob{K}{\textbf{D}}\cprob{J}{K}\cprob{\textbf{X}}{\textrm{pa}(\textbf{X})}}{\cprob{J}{K}\cprob{\textbf{X}}{\textrm{pa}(\textbf{X}} \int_{\textbf{D}} \uprob{\textbf{D}}\cprob{K}{\textbf{D}}}\\  
            &= \cprob{\textbf{D}}{K}.
        \end{align*}
        $\cprob{\textbf{D}}{K}$ is available in closed form as the Dirichlet distribution is conjugate to the Categorical distribution. Access to hidden variables during training simplifies posterior inference considerably, because Markov independence properties allow us to drop otherwise intractable densities (e.g., $\cprob{\textbf{X}}{\textrm{pa}(\textbf{X}}$). 
        
        In training we start with uninformed priors and let the model evolve during training according to the following selection strategy for $\mathcal{S}$. Using a pre-selected subset of validation samples, we build $\mathcal{S}$ from those samples, that exhibit a high loss according to the objective function of training. Over multiple epochs, this procedure tunes the prior probability distributions to better match the expectations of the real world. We emphasize that, depending on the task at hand, other selection strategies, such as measuring the uncertainty of perturbated inputs \cite{gal2015dropout}, may also be applicable.
        
        

\section{Experiments}
\begin{figure}[htbp]
            \centering
            \iffinalcopy
                \includegraphics[width=\columnwidth]{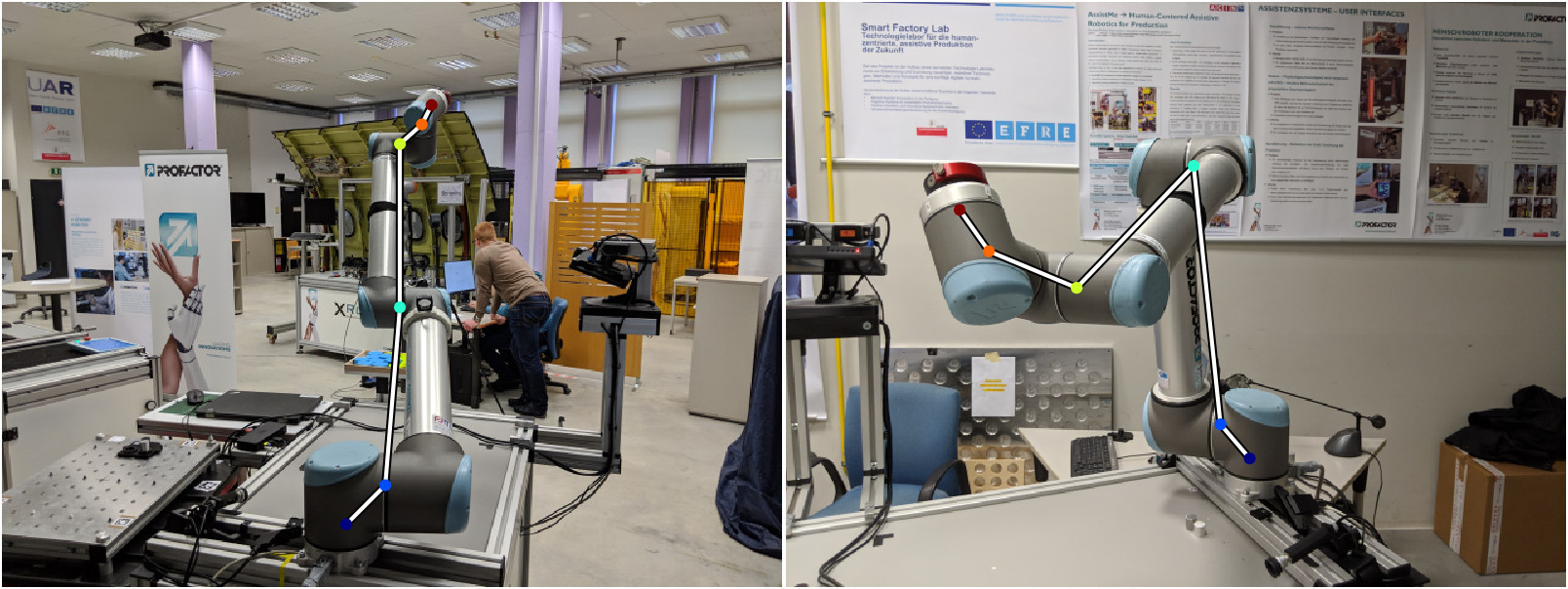}
            \else
                \includegraphics[width=0.85\columnwidth]{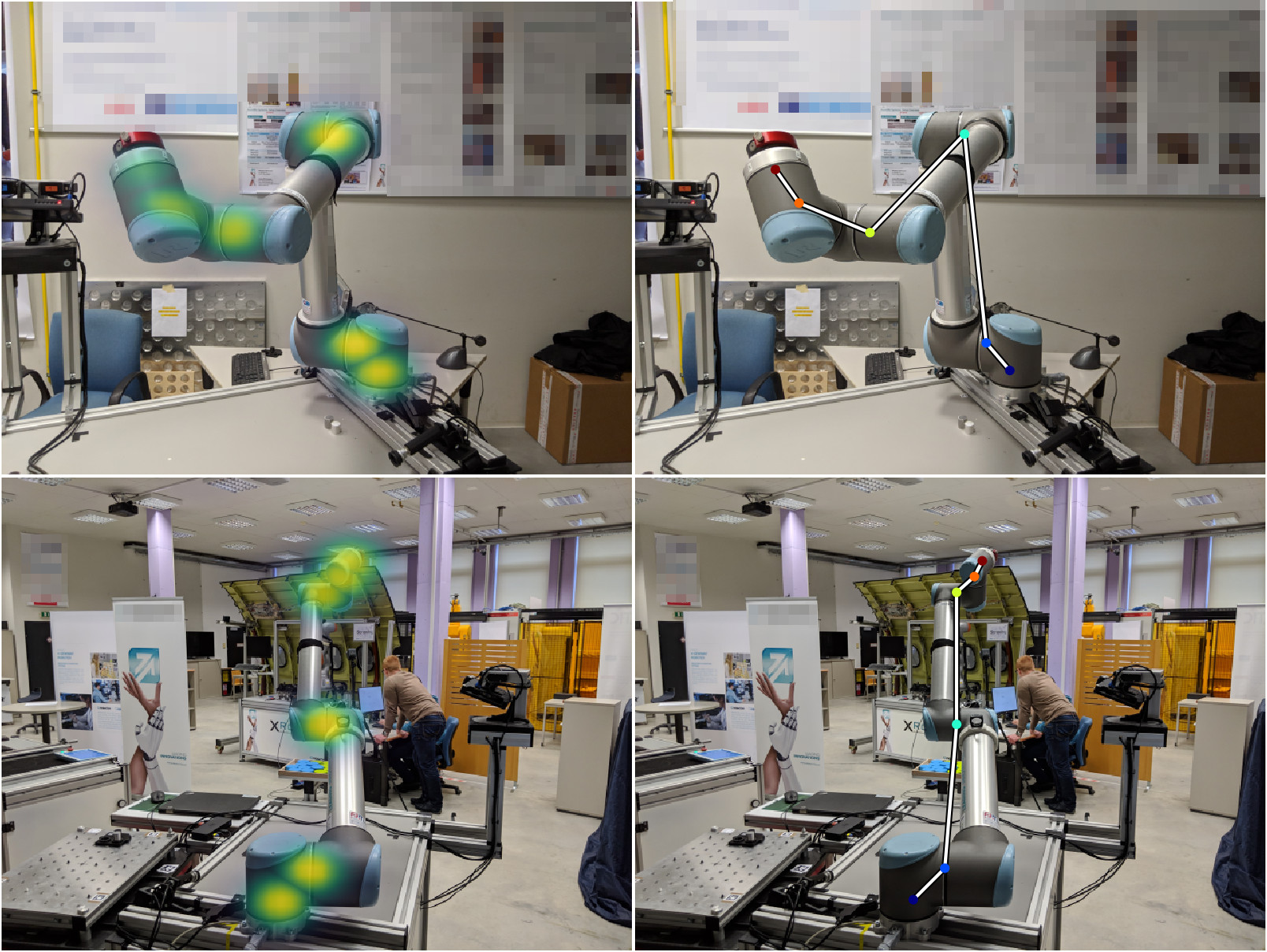}
            \fi
            \caption{
                \label{fig:posresults} Qualitative real-world database results.
            }
        \end{figure}

    We evaluate the accuracy of the localization network, trained on artificial views of an UR10 robot model (see \figurename \ref{fig:posresults}). We report results for synthetic and real-world images and compare these to human-level performance. Finally, we study the effects of the proposed feedback mechanism on training progress.

    \textbf{Training setup} 
    We use Blender as a simulator engine for robotic scenes and PyTorch to train the localization model. For distributed communication, we build upon ZeroMQ and rely on publisher/subscriber pattern for scalability \footnote{Code available at \iffinalcopy \url{https://github.com/cheind/pytorch-blender} \else \texttt{Link available upon acceptance.} \fi}. 
        
    To train the localization model we sample a set of 10,000 examples per epoch from up to eight simulator engines plus an additional set of 2,000 images for validation. We initialize each simulator with uniform random distributions. Render resolution is set to $640 \times 480$. Additional augmentation includes random backgrounds (city and industrial themes), image distortions and random distractors (lines and other primitives). Additionally, we gathered and annotated 100 real-world images of a UR10 robot for validation purposes only. 

    
    \textbf{Metrics} We define the localization error as the Euclidean distance between predicted and target pixel joint locations and report it in percent of the image diagonal. To extract joint locations from belief maps $\hat{\mathbf{b}}$, we proceed as follows: first, we apply non-maximum suppression and then we extract the maximum peak per joint.

    \subsection{Localization Accuracy and Human Annotation Uncertainty}       
\begin{figure}[!h]
            \centering
            \begin{subfigure}[t]{0.49\columnwidth}
                \includegraphics[width=\columnwidth]{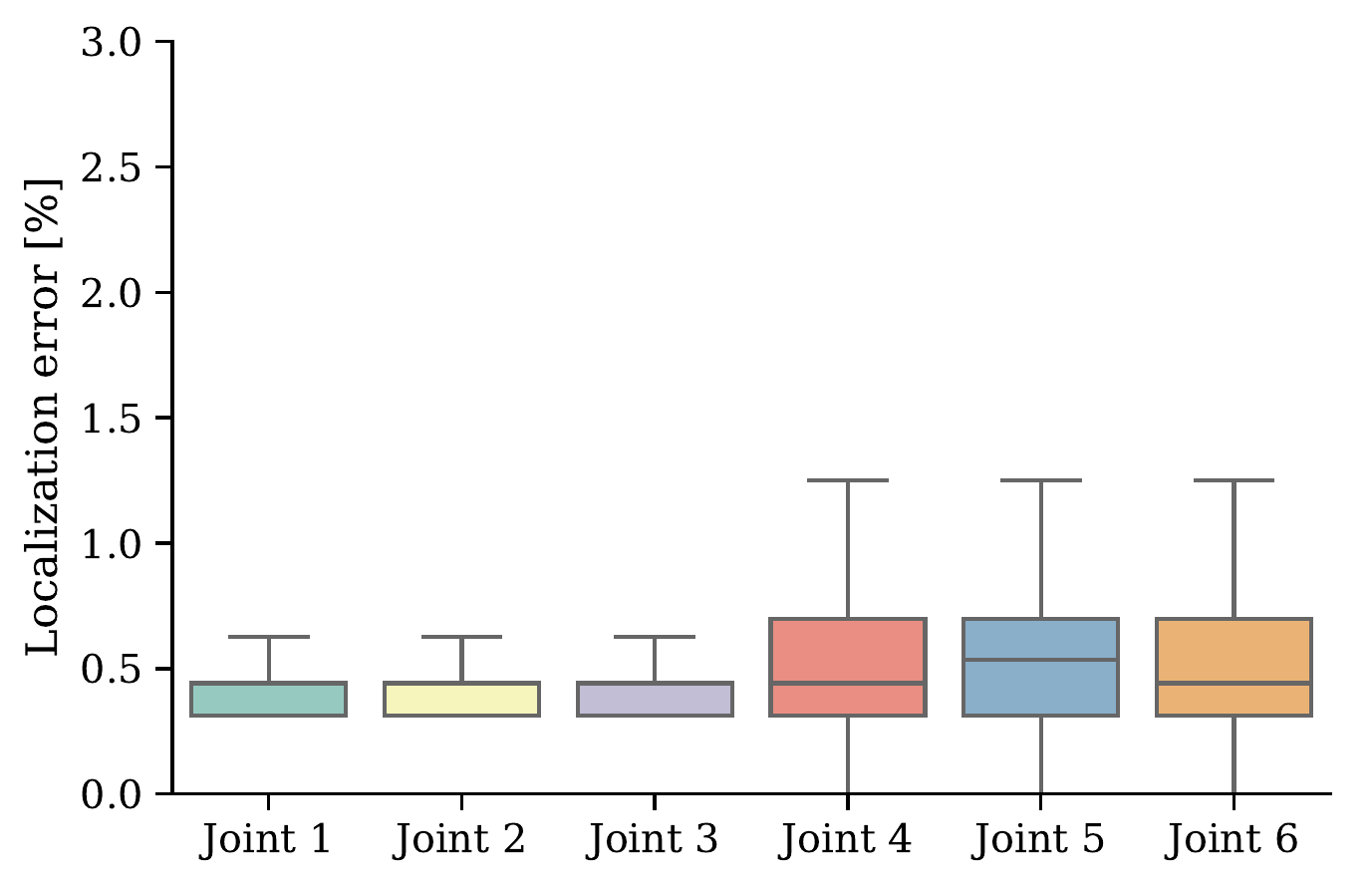}
                \caption {
                    \label{fig:locationerror_synth} 
                    Localization errors of joint prediction on a synthetic test data set with auto-labeled ground truth.
                }
            \end{subfigure}
            \begin{subfigure}[t]{0.49\columnwidth}
                \includegraphics[width=\columnwidth]{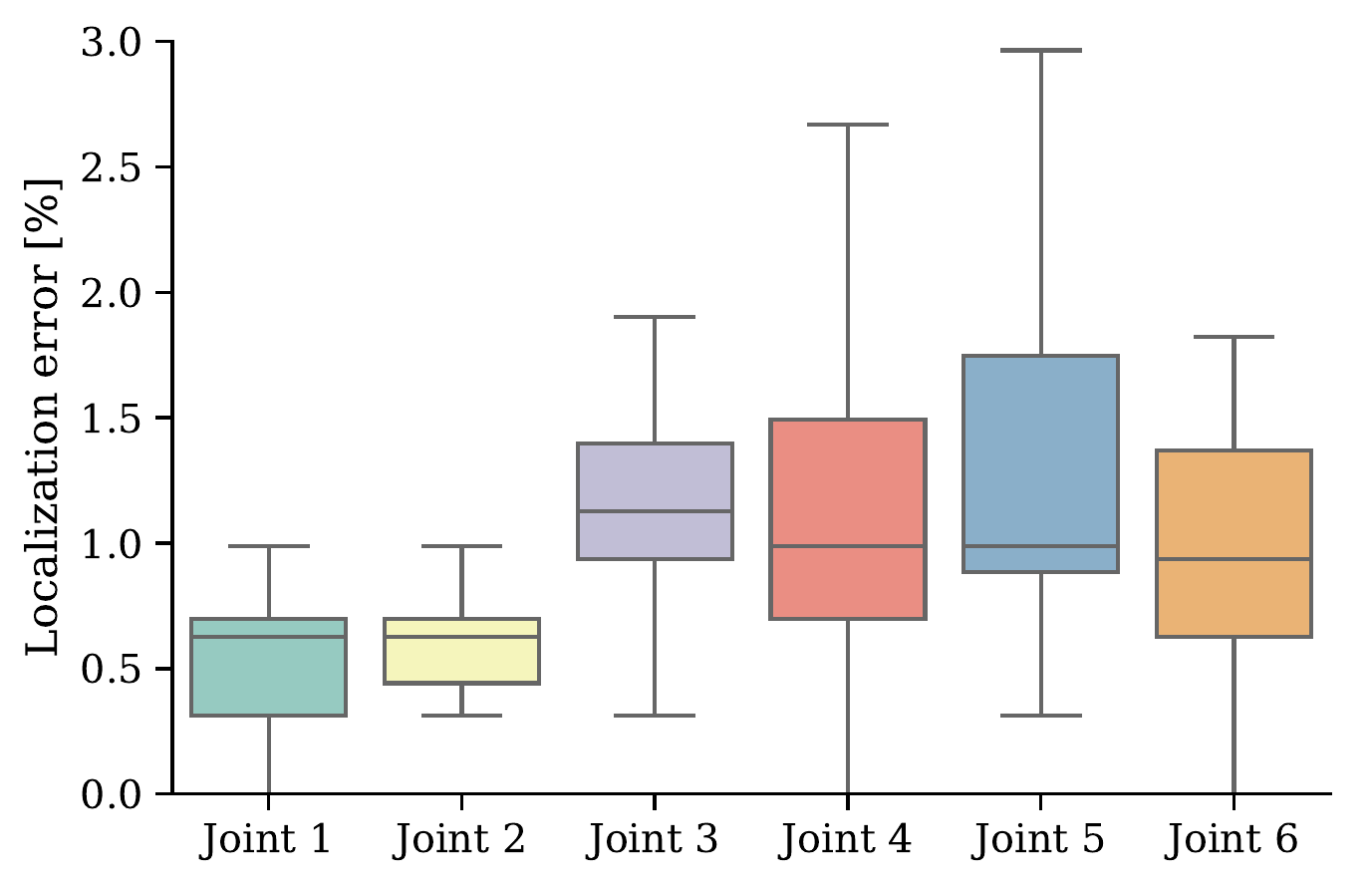}
                \caption {
                    \label{fig:locationerror_real} 
                    Localization errors of joint prediction for a real-world test set. Target locations annotated by a human domain expert. 
                }
            \end{subfigure}
            \caption {
                \label{fig:pixelerror} 
                Localization errors of robot joint prediction. Results are in percent of image diagonal.
            }
        \end{figure}

        \figurename \ref{fig:pixelerror} reports localization errors on two distinct test sets: synthetic with random backgrounds (\figurename \ref{fig:locationerror_synth}) and real-world images (\figurename \ref{fig:locationerror_real}). Our system achieves a mean localization error of \SI{0.6}{\percent} on synthetic and \SI{1.3}{\percent} on real data, averaged over all robot joints.

        We attribute a slight loss in performance between synthetic and real data to two reasons: (a) our simulation does not capture all important aspects of real-world images and (b) human annotation is error-prone. 
        
\begin{figure}[!h]
            \centering
            \begin{subfigure}[t]{0.49\columnwidth}
                \includegraphics[width=\columnwidth]{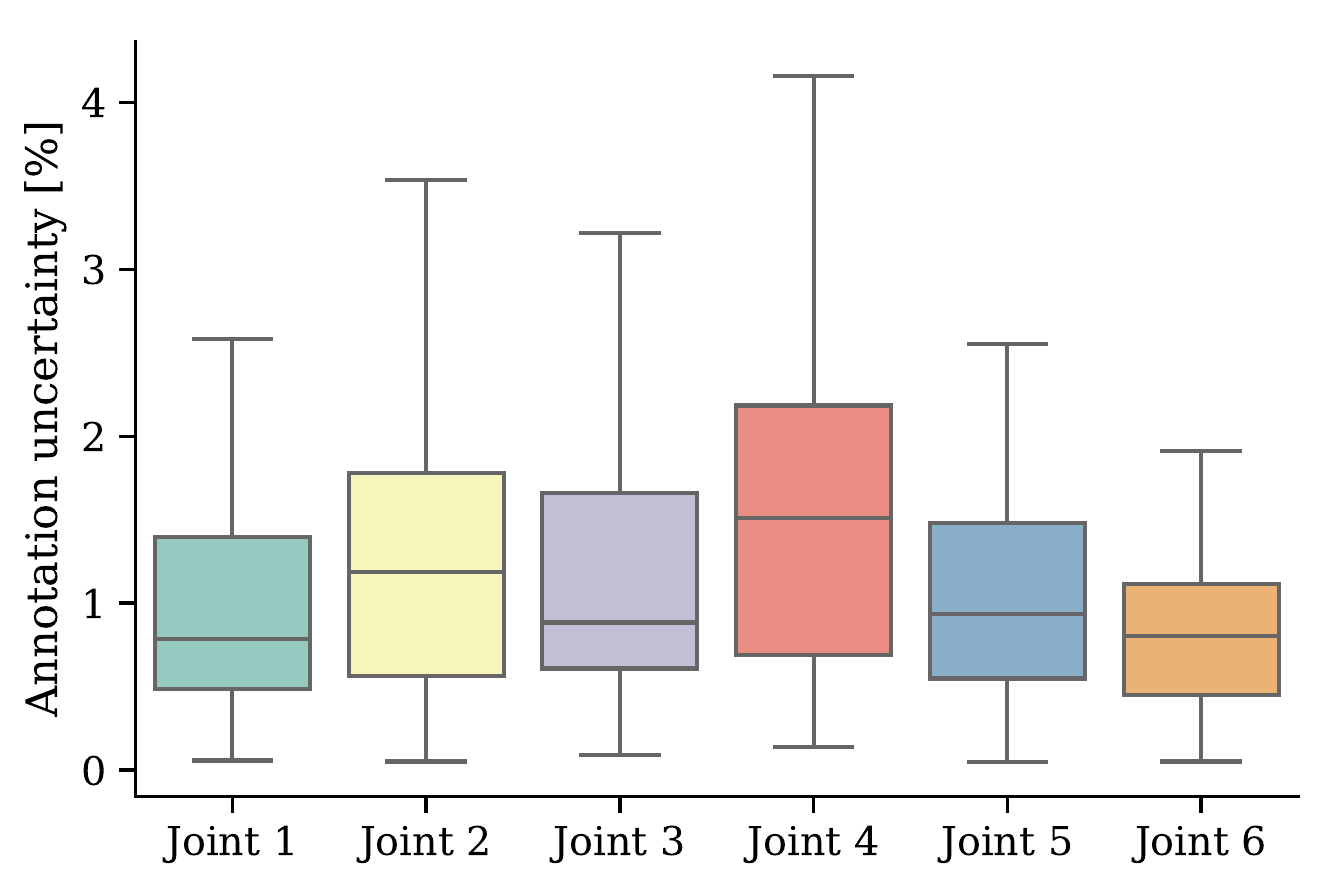}
                \caption {
                    \label{fig:humanuncertainty} 
                    Human uncertainty estimates in keypoint annotation. Results are in percent of image diagonal.
                }
            \end{subfigure}
            \begin{subfigure}[t]{0.49\columnwidth}
                \iffinalcopy
                    \includegraphics[width=\columnwidth]{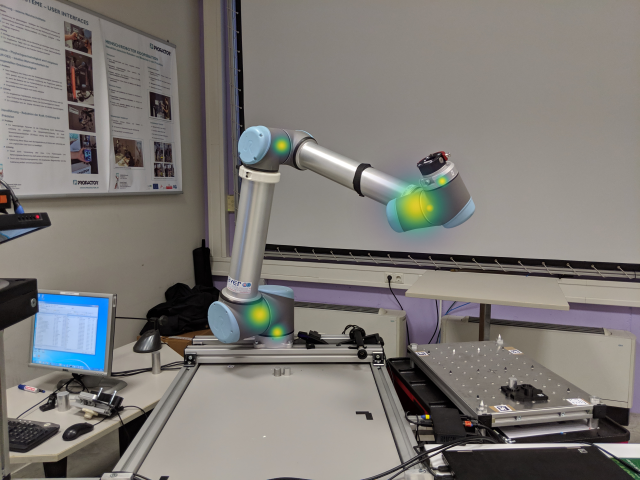}
                \else
                    \includegraphics[width=\columnwidth]{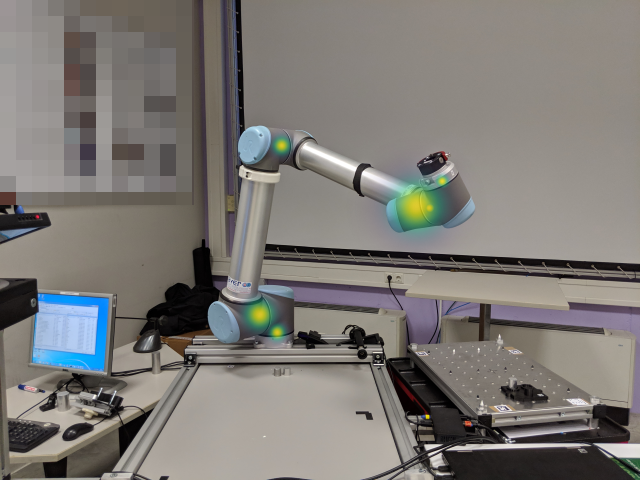}
                \fi
                \caption {
                    \label{fig:humanuncertainty_belief} 
                    Uncertainty estimates superimposed as Gaussian kernels on one of the images to be annotated.
                }
            \end{subfigure}
            \caption {
                \label{fig:uncertainty} 
                Uncertainty estimates from human annotation. 
            }
        \end{figure}

        To substantiate the latter, we conducted a pilot study that asked 10 robotic experts to annotate a set of 12 real-world images of an UR-10 robot. \figurename \ref{fig:humanuncertainty} shows the inter-rater localization error spread per joint over all images in percent of the image diagonal. We find that the middle joints, often not visible because of self-occlusion, are the most difficult to annotate precisely. These results indicate that the network's performance is similar to human-level performance on the given images.

    \subsection{Effects of Feedback}
\begin{figure}[htbp]
            \centering
            \includegraphics[width=0.9\columnwidth]{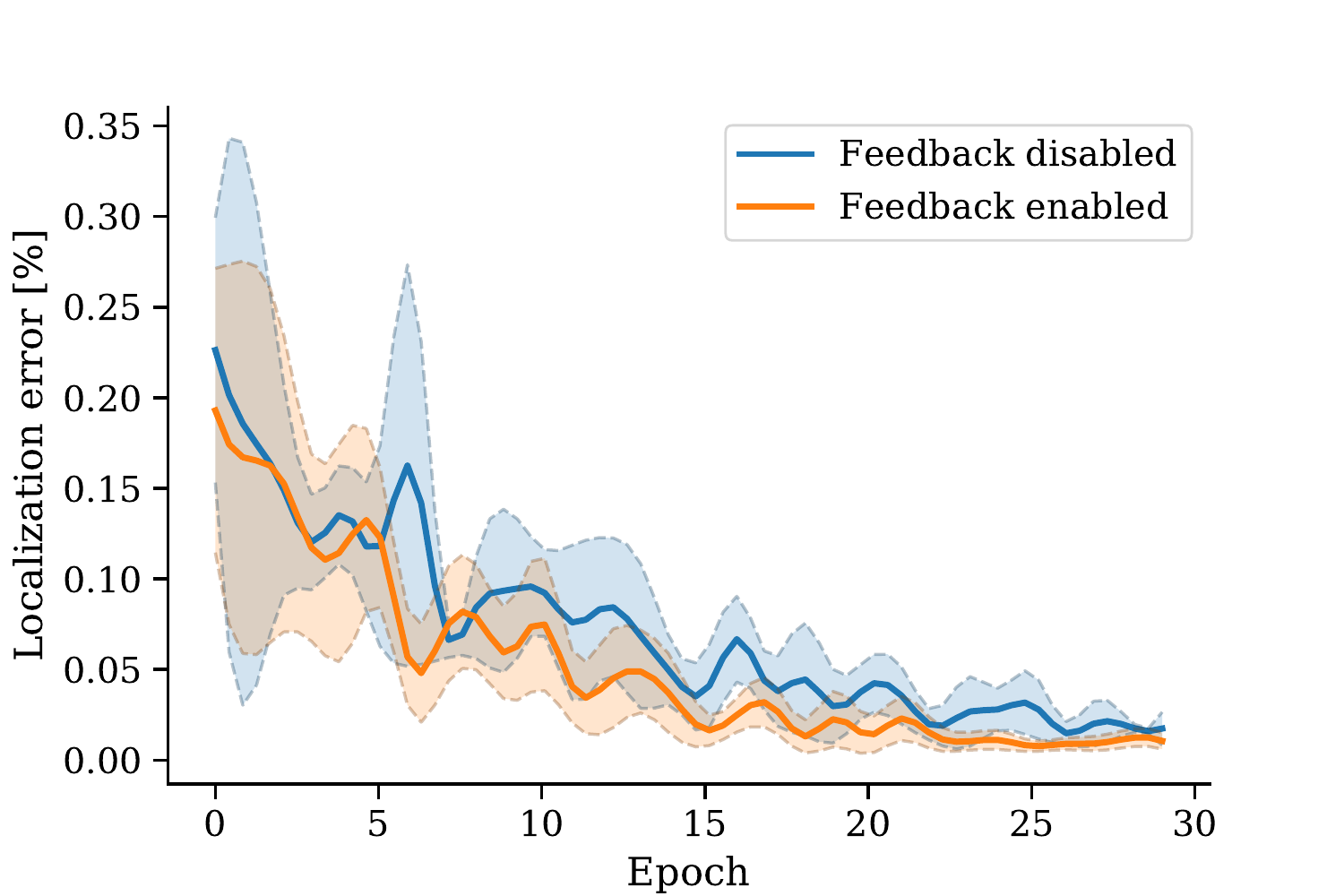}
            \caption{
                \label{fig:control} Mean and 1-$\sigma$ standard deviation of localization errors over 500 trainings, comparing feedback-enabled and feedback-disabled trainings. We see faster convergence if we let one of the simulator instances generate more examples in areas of high training error. Errors are in percent of image diagonal.
            }
        \end{figure}

        We evaluate the efficiency of our feedback mechanism as follows: we perform 500 trainings, each training is run twice. Feedback is disabled in the first run and enabled in the second run. Each run spawns two simulator engines. In the feedback-enabled run, we adapt the prior distributions over joint angles of the second simulator to generate more samples associated with high validation error as outlined in sub-section \ref{sec:adaptivetraining}. \figurename \ref{fig:control} shows the beneficial effects of our approach. Our findings indicate that the same level of accuracy can be reached in fewer iterations when feedback is enabled.

    \subsection{Runtime Analysis}
        Table \ref{tab:runtime} summarizes the runtimes of our system (Intel 12-Core and 4 NVIDIA Tesla V100 GPUs). In terms of performance, our framework generates samples quickly enough to prevent the model training loop from stalling. The simulation timings include sampling from the probabilistic model, image rendering, sample gathering. A single train step includes timings for forward and backward model passes as well as computing and applying feedback. In all experiments the training batch-size was set to eight. The outputs of different simulators are interleaved during batch composition, ensuring fair sample processing.

        \begin{table}[htp]
            \centering
            \begin{tabular}{lrr}
                \toprule
                {} &  Mean [ms] &  Std [ms] \\
                Task $640 \times 480$  &            &           \\
                \midrule
                Simulation step        &      321.23 &    24.76  \\
                Train step             &      521.23 &   73.76  \\
                \bottomrule
            \end{tabular}
            \caption{
                \label{tab:runtime} 
                Runtimes of different parts of our pipeline. When performed in parallel (8 instances), sampling from the model is faster than a single training step.
            }
        \end{table}

\section{Conclusion and Discussion}
We have described a learning based approach for the prediction of robot joint positions in color images. The proposed architecture trains solely from probabilistically generated computer images and offers desirable properties, including scalability and adaptive data sampling. The presented results show the method's ability to generalize predictions to real world photos with an accuracy that comes close to human-level performance.

Our approach has a number of shortcomings in its current state. Thus far, we have trained with only one type of robot and have not considered multiple robot instances. For an efficient inference, the proposed feedback mechanism requires access to hidden random variables, which may not always be possible depending on the application. In the future, we plan to apply our methods to other robot types and explore the effects of the proposed feedback mechanism in other domains. 

\pagebreak

\iffinalcopy
\section*{Acknowledgment}
This research was supported by ”FTI Struktur Land Oberoesterreich (2017-2020)”, the EU in cooperation with the State of Upper Austria within the project Investition in Wachstum und Besch\"aftigung.
\else
\fi

\small
\bibliographystyle{ieeetr}
\bibliography{biblio}

\end{document}